\documentclass[runningheads]{llncs}
\pdfoutput=1
\usepackage{graphicx}

\usepackage{subfigure}
\usepackage{caption}
\usepackage[linesnumbered, ruled]{algorithm2e}

\begin{document}
\title{Hierarchical Genetic Algorithms with evolving objective functions\thanks{Code available at \texttt{https://github.com/Harsha061/CMC\_Proj}}}
%
%
\author{Harshavardhan P K\inst{1} \and
Kousik Krishnan\inst{2}}
\authorrunning{Harshavardhan, Krishnan}
%
\institute{Dept. of Computer Science and Engineering\\ Indian Institute of Technology Madras\\
\email{harshavardhan864.hk@gmail.com}\and
Chennai Mathematical Institute\\
\email{kousik@cmi.ac.in}\\
}
\maketitle              
\begin{abstract}
We propose a framework of genetic algorithms which use multi-level hierarchies to solve an optimization problem by searching over the space of simpler objective functions. We solve a variant of Travelling Salesman Problem called \texttt{soft-TSP} and show that when the constraints on the overall objective function is changed the algorithm adapts to churn out solutions for changed objective.

We use this idea to speedup learning by systematically altering the constraints to find a more globally optimal solution.

We also use this framework to solve polynomial regression where the actual objective function is unknown but searching over space of available objective functions yields a good approximate solution.

\end{abstract}
\section{Introduction}
Genetic Algorithms(GAs) are used in many global optimization problems like travelling salesman problem(TSP), protein structure prediction\cite{pedersen1996genetic,krasnogor1999protein,unger2004genetic}, design problems\cite{morris1998automated}, data fitting\cite{karr1995least}, phylogenetics \cite{zwickl2006genetic}, etc.

The crux of GAs are to search for right set of parameters to optimize a fixed objective function. Can we learn faster if we evolve objective functions such that the solutions reach better optimum or converge faster? We introduce the notion of hierarchy over objective functions where lower level objective functions continuously adapt to solve the higher level objective. These lower level objectives can be a constrained versions of higher level objectives so that they search in specific areas of solution space.

Alternatively, lower level objectives can be a set of good approximations to the more complex or obscure higher level objective.

We leverage these ideas to propose a hierarchical framework for GAs. We then solve variation of travelling salesman problem called \texttt{soft-TSP} and problem of polynomial regression where objective function is unclear.

\section{Preliminaries}
\subsection{Genetic Algorithms}
Genetic Algorithms (GA) are heuristics used for search. They are inspired from phenomenon of evolution, in particular the mechanism of \emph{Natural Selection} first proposed by Charles Darwin in his seminal work \emph{Origin of Species} \cite{darwin2004origin}. Natural Selection essentially says that the basis of survival of species is that the individuals with favourable genetic expressions are selected over the unfavourable ones to pass their genetic information to the next generation of off-springs.

In genetic algorithms(GA), each \emph{individual} is a solution to the problem. An individual is defined by its \emph{genotype} which are sequence of \emph{genes}. The location of a gene is called \emph{locus}.

Each \emph{generation} $G$ has a set of individuals. A \emph{fitness function} $F:G\to \mathbf{R}$ is defined to measure how good a solution is. GAs consisit of three components (or operations) \cite{goldberg2006genetic}:
\begin{enumerate}
    \item \textbf{Reproduction}: Some of the individuals in current generation are copied on to next generation. A \emph{selection strategy} is used such that the more \emph{fit} individuals have higher chances of being copied to next generation.
    
    \item \textbf{Crossover:} In this operation, two individuals' \emph{genotype} is combined to produce a new \emph{offspring} that shares parts of genotype of both parents.
    
    \item \textbf{Mutation:} Mutation alters the genotype of individual randomly to produce a slightly different genotype.
\end{enumerate}

Using reproduction and crossover may narrow down the search space over time and we may lose opportunity to explore some important areas of search space prematurely. Hence, mutation allows for random exploration and helps maintain the diversity of solutions in each generation.

Rate at which crossover and mutation is done are some hyperparameters of the GA. The flow of a GA is summarized in Algorithm \ref{alg:ga}.

\begin{algorithm}[h]
\SetKwInOut{Input}{input}
\SetKwInOut{Output}{output}
\SetAlgoLined
$gen=0$\;
Initialize population\;
\Repeat{until convergence}{
    Evaluate fitness of individuals\;
    Select individuals for next generation\;
    Apply crossover to pairs of chosen individuals to get new off-springs\;
    Mutate some of the individuals\;
}
Output the best solution based on fitness\;
\caption{Genetic Algorithm}
\label{alg:ga}
\end{algorithm}

\subsubsection{Selection Strategies for GA}
We experimented with following selection strategies.
\paragraph{Random selection} We select fraction of individuals randomly with replacement for next population. It is not a good strategy, and is used only for benchmarking and debugging.

\paragraph{Softmax Selection} Select based on softmax probabilities of its fitness.

\paragraph{Percentile selection} Select top $p$ based on fitness function.

\paragraph{Random tournaments} This strategy selects one individual in each round. We repeat for sufficient number of rounds.
At each round, we randomly sample $k$ individuals and choose the best among them and don't consider it for further rounds.

After selection, each individual is crossed with another random individual with probability $c'$ and offspring is added to population. Then, the individual is mutated and new mutated individual is added.

Some of the hyperparameters to tune are:
\begin{itemize}
    \item \textbf{Crossover rate} $c'$: Probability of individual getting crossed with a random individual.
    \item \textbf{Point crossover probability} $c$: Probability that a gene value is swapped during crossover.(More details are provided when solving specific problems)
    \item \textbf{Mutation rate} $m$: Probability that a specific gene is mutated during mutation.
\end{itemize}

\subsection{\texttt{TSP} and \texttt{soft-TSP}}
A weighted graph is represented by $G=(V,E,W)$ where $V$ is set of vertices, $E\in V\times V$ the set of undirected edges and $W:E\to \mathbf{R}$ a weight function on edges. 
\begin{definition}
A \emph{path} $P=(V',E'), V'\subseteq V, E'\subseteq E$ of graph $G$ is a series of vertices $V'=[v_1,v_2,\dots,v_m]$ such that $E'=\{(v_i,v_{i+1})\in E|  i\in \{1,2,\dots,m-1\}\}$.

\end{definition}

\begin{definition}
A \emph{Hamiltonian Path} $P=(V',E')$ is path that contains all vertices exactly once. The \emph{cost} of a Hamiltonian path, $w(P)$, is sum of weights of all edges in the path.
\[w(P)=\sum_{e\in E'}w(e)\]
\end{definition}
\begin{definition}
A \emph{Complete Graph} has an edge between any pair of vertices.
\end{definition}
Thus any permutation of vertices in complete path forms a Hamiltonian path.

In travelling salesman problem (\texttt{TSP}) we are given a complete graph and we need to find a Hamiltonian path of least weight. This problem is shown to be NP-Hard \cite{karp1975computational}.

The \texttt{soft-TSP} is a variant of \texttt{TSP} problem where each vertex has a penalty $p:V\to \mathbf{R}$.A feasible solution of \texttt{soft-TSP} is Hamiltonian path $P'$ of a subgraph $G'(V',E')$ of $G$, where $V'\subseteq V$ and the cost is $\sum_{v\not\in V'}p(v)+w(P')$. The cost of Hamiltonian path is summed with the penalties of vertices not considered in the path. Clearly, \texttt{soft-TSP} is harder than \texttt{TSP}.

A good practical example for this problem is applicable when we are planning to go on a trip and visit various cities. The importance of visiting a city is encoded by its penalty. The solution is the order in which we visit the cities such that least important ones can be missed while reducing the total distance of the trip.

\subsection{Objective functions for Regression}
In regression task we are given $X=\{x_1,x_2,\dots,x_N\}$ and $Y=\{y_1,y_2,\dots,y_n\}$ where $\forall i:y_i=p(x_i)+\epsilon$ for some function $p$ and noise $\epsilon$. We need to approximate this function by finding a function $p'$ and $y'_{i}=p'(x_i)$.

To measure how well $p'$ approximates $p$, we choose an objective function to minimize. There are many objective functions for regression task.

\begin{figure}[h]
    \centering
    \includegraphics[width=1.2\textwidth]{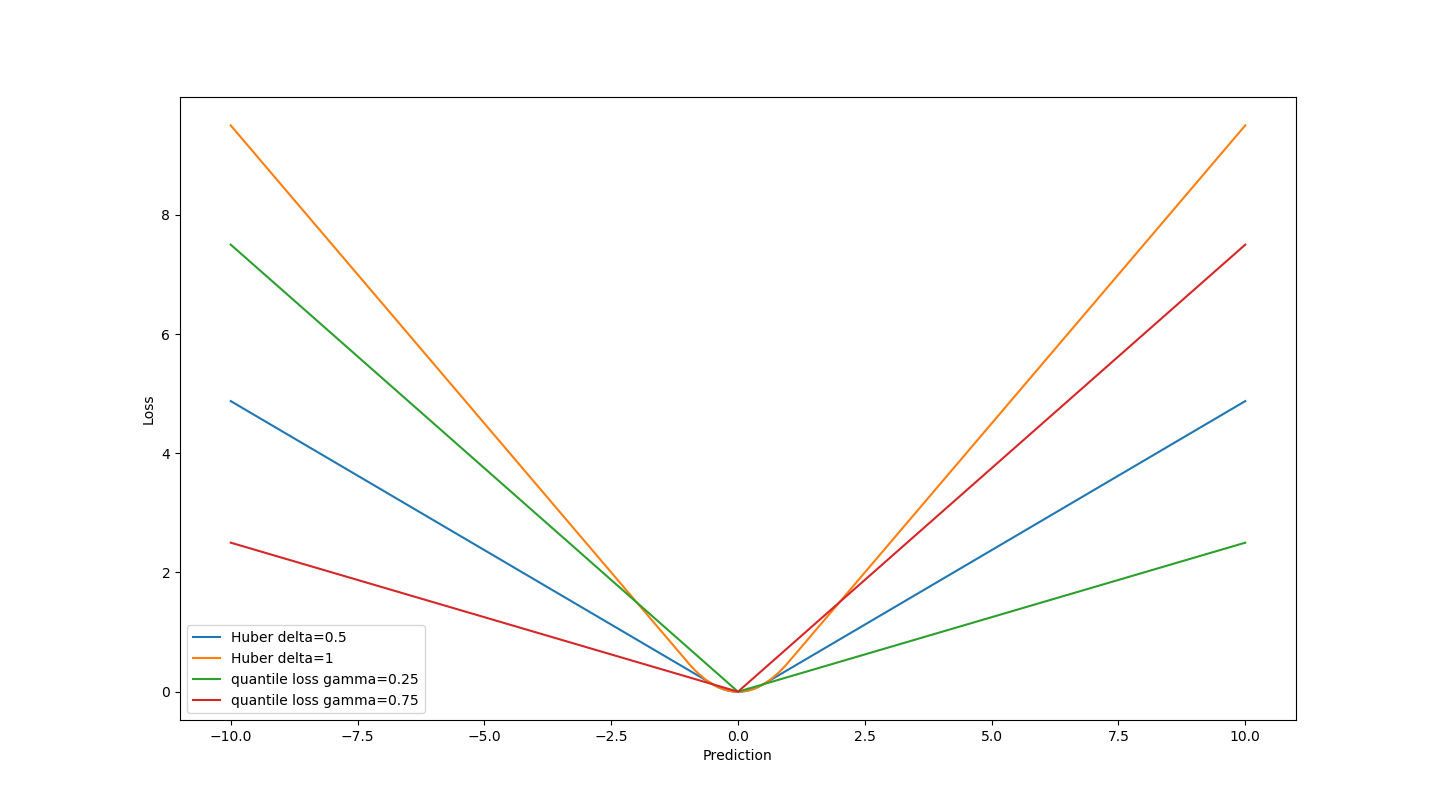}
    \caption{Different values of loss function based on bias from ground truth}
    \label{fig:loss}
\end{figure}

\begin{itemize}
    \item \textbf{MSE}: Mean squared error loss. This is the most commonly used objective function which is easily differentiable.
    
    \[\frac{1}{2N}\sum_{i=1}^N(p(x_i)-p'(x_i))^2\]
    \item \textbf{MAE}: Mean Absolute Error
    \[\frac{1}{N}\sum_{i=1}^N|p(x_i)-p'(x_i)|\]
    \item \textbf{Quantile Loss:} It implements different weights to positive and negative bias.
    \[\frac{1}{N}(\sum_{p(x)>p'(x)}\gamma|p(x)-p'(x)|+\sum_{p(x)<p'(x)}(1-\gamma)|p(x)-p'(x)|)\]
    \item \textbf{Huber Loss:} It is a combination of MSE and MAE.
    \begin{center}
        \includegraphics[width=.8\textwidth]{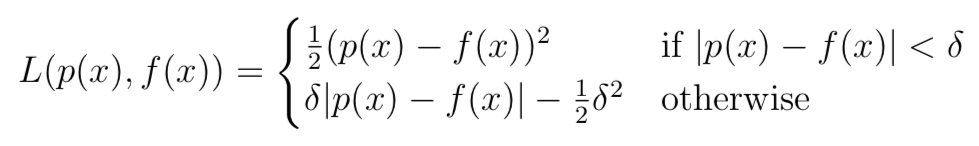}
    \end{center}
    \item We can have many other variations such as having weights associated with the range of value of $x$ that denote which range to focus on.
\end{itemize}
The loss functions are visualized in Figure \ref{fig:loss}.

\section{Relevant Works}

The work by Bentley et.al \cite{bentley1996hierarchical} presented a novel method of crossover(as compared to traditional method of cross-over by uniform selection of parameters to exchange), where they use domain-knowledge to construct a \emph{Semantic tree of genotype} to enable separate cross-overs for different sub-groups of genes. 

The Work by Sobey et.al \cite{sobey2018re} proposes to model individuals as belonging to different populations having  different population-specific fitness functions and a common global fitness function. This could be used to explore solutions from different constrained sub-spaces. They propose different selection strategies across populations and within population. However, unlike our framework their population specific fitness functions are pre-determined and fixed.

 There are also few selection strategies for TSP problem as discussed in \cite{braun1990solving,razali2011genetic}. Some of these strategies are incorporated in our work.
\section{Hierarchical GA framework}

\subsection{Motivation}
Consider the problem of solving a jigsaw puzzle. We tend to first get partial solutions for different areas of the puzzle and then combine them together to get complete solution. In other words, we use simpler sub-problems to come up with full, more complex solutions. Moreover, the way we choose these simpler sub-problems are also evolving. We try out different regions, get better ideas about what sub-regions map to which pieces and formulate better sub-problems to solve and explore jigsaw space more fully.
\begin{figure}[h]
    \centering
    \includegraphics[width=0.6\textwidth]{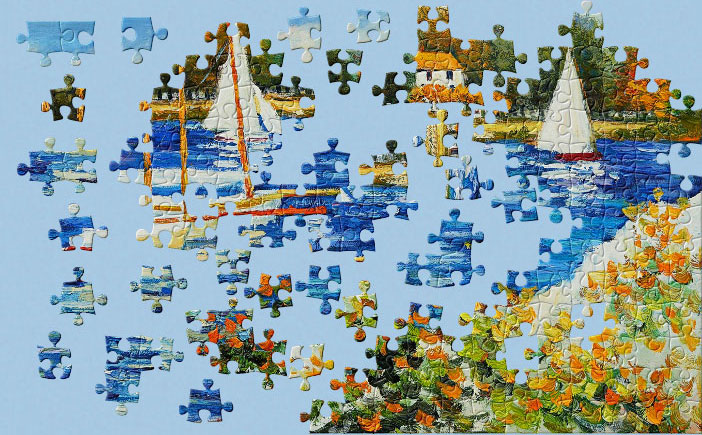}
    \caption{Partial solutions of Jigsaw puzzle}
    \label{fig:my_label}
\end{figure}
\subsection{Setup}
One way to view this is by considering this process as carefully evolving and adapting the space of sub-problems to solve so that we get closer to solving the complete problem. This produces notion of hierarchy of jigsaw solvers, with the root solver (trying to solve the global problem) controlling the way the solvers evolve to get a better overall solution. This is similar to the some frameworks used in hierarchical reinforcement learning.\cite{barto2003recent}

We can extend this notion to Genetic Algorithms.
Consider an optimization problem $P$ and a family of problems $F$ such that feasible solution for a problem in $F$ is a feasible solution for $P$. If the optimal solution of $P$ is also an optimal solution in some of the problems in $F$, then searching over space of $F$ by solving problems in $F$ can be a method to find optimal solution for $P$, especially if $F$ is much easier to solve.

The \emph{meta-solver} aims to solve $P$.
The \emph{meta-solver} spawns individuals which are \emph{GA solvers} for simpler problems in $F$. The \emph{GA solvers} in-turn solve respective sub-problems in $F$. The meta-solver controls how the sub-problem solvers \emph{evolve}. The \emph{meta-solver} searches in the space of sub-problems by evolving the population of sub-problems to find the best solution. The procedure is summarized in Algorithm \ref{alg:hga}.

\begin{figure}[h]
    \centering
    \includegraphics[width=\textwidth]{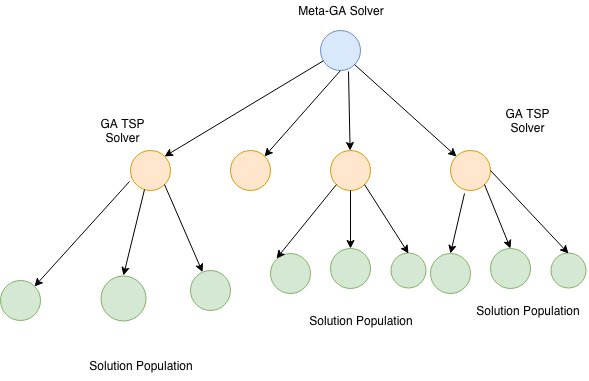}
    \caption{Hierarchical GA Framework}
    \label{fig:my_label}
\end{figure}

\begin{algorithm}[h]
\SetKwInOut{Input}{input}
\SetKwInOut{Output}{output}
\SetAlgoLined
Initialize meta-solver with individuals in $I$\;
\Repeat{until convergence}{
    Train each of the sub-solvers in $I$ for $k$ generations\;
    Evolve population $I$\;
}
\caption{Training algorithm for Hierarchical GA Framework}
\label{alg:hga}
\end{algorithm}

\section{Solving \texttt{soft-TSP}}
\texttt{soft-TSP} problem involves first finding the right subgraph of graph $G=(V,E)$ on which TSP in computed. The \emph{meta-solver} searches over space of subsets of vertex set while \emph{GA solvers} solve the \texttt{TSP} problem for given subset of vertex.

Here, we consider only complete graphs and vertices are points in two-dimensional Euclidean space with weights of edges equal to the euclidean distance.
\subsection{Details of meta-solver}
\paragraph{Genotype} : Each individual $i$, in the \emph{meta-solver} population, is a boolean array $B_i$ of size $n=|V|$. If $B_i[j]=1$, vertex $v_j$ is part of subset assigned to $i$.

\paragraph{Crossover} : The algorithm for crossover between two individuals is described in Algorithm \ref{alg:co_tsp}. In brief, some of the values of boolean arrays are swapped.

\begin{algorithm}[h]
\SetKwInOut{Input}{input}
\SetKwInOut{Output}{output}
\SetAlgoLined
\Input{2 boolean arrays $B_1,B_2$, point crossover rate $c\in[0,1]$}
\For{$i\leftarrow \{1,\dots,|V|\}$}{
Select $r\leftarrow[0,1]$ uniformly at random\;
\If{$r<c$}{
    $swap(B_1[i],B_2[i])$\;
}
}
return $B_1,B_2$\;
\caption{Crossover for meta-solver of soft-TSP}
\label{alg:co_tsp}
\end{algorithm}

\paragraph{Mutation} :  The algorithm for mutation  is described in Algorithm \ref{alg:mu_tsp}. In brief, some of the values of boolean array are flipped.

\begin{algorithm}[h]
\SetKwInOut{Input}{input}
\SetKwInOut{Output}{output}
\SetAlgoLined
\Input{boolean arrays $B$, mutation rate $m\in[0,1]$}
\For{$i\leftarrow \{1,\dots,|V|\}$}{
Select $r\leftarrow[0,1]$ uniformly at random\;
\If{$r<m$}{
    $B[i]\leftarrow 1-B[i]$\;
}
}
return $B$\;
\caption{Mutation for meta-solver of soft-TSP}
\label{alg:mu_tsp}
\end{algorithm}

\paragraph{Selection Strategy} : We use \emph{Percentile selection} with percentile $p=50$. Then we do crossover and mutation as described at end of Section 2.1.

Also, if the population size decreases to less than $20\%$ of population size of first generation, we crossover the best solution with another solution chosen randomly from selected individuals to produce new individuals. Then, we mutate the new individuals before adding to the population.

\begin{table}[h]
    \caption{Hyperparameters for meta-Solver of \texttt{soft-TSP}}
    \label{tab:mtsp}
    \centering
    \begin{tabular}{|c|c|}
    \hline
    \textbf{Hyperparameter} & \textbf{Value}\\
    \hline
     Initial Population Size    &  100\\
     Minimum population Size    & 20\\
     Mutation Rate & 0.2\\
     Crossover rate & 0.5\\
     Point crossover probability & 0.5\\
     Number of generations of sub-solvers for one generation of meta-solver & 50\\
         \hline
    \end{tabular}

\end{table}

\subsection{Details of \texttt{TSP} solver}
We now describe details of \texttt{TSP} solver used in second level of hierarchy. The \texttt{TSP} solvers are given the vertices on which Hamiltonian path is to be computed. As mentioned earlier, any permutation of these vertices gives a valid Hamiltonian path since graph is complete.

\paragraph{Genotype} Each individual $i$ has an array $P_i$ which is a permutation on the vertices.

\paragraph{Crossover} The crossover strategy used was from \cite{braun1990solving} and is described in Algorithm \ref{alg:co_tsp2}.

\begin{algorithm}[h]
\SetKwInOut{Input}{input}
\SetKwInOut{Output}{output}
\SetAlgoLined
\Input{2 arrays $P_1,P_2$, point crossover rate $c\in[0,1]$}
Toss a coin of bias $c$ and record the results in boolean array $B[1\dots,|V'|]$\;
Initialize empty lists $L_1,L_2$
\For{$i\leftarrow \{1,\dots,|V'|\}$}{
\If{$B[i]=1$}{
Append $P_1[i]$ to $L_1$\;
}
}
\For{$i\leftarrow \{1,\dots,|V'|\}$}{
\If{$P_2[i]$ is not in $L_1$}{
Append $P_2[i]$ to $L_2$\;
}
}
Concatenate $L1$ and $L2$ to get $P_3$\;
return $P_3$\;
\caption{Crossover for TSP solver}
\label{alg:co_tsp2}
\end{algorithm}

\paragraph{Mutation} The algorithm for mutation  is described in Algorithm \ref{alg:mu_tsp2}. In brief, some of the values permutation array are swapped.

\begin{algorithm}[h]
\SetKwInOut{Input}{input}
\SetKwInOut{Output}{output}
\SetAlgoLined
\Input{arrays $P$, mutation rate $m\in[0,1]$}
\For{$i\leftarrow \{1,\dots,|V|\}$}{
Select $r\leftarrow[0,1]$ uniformly at random\;
\If{$r<m$}{
    Select $j\leftarrow V'$ at random\;
    $swap(P[i],P[j])$\;
}
}
return $P$\;
\caption{Mutation for TSP solver}
\label{alg:mu_tsp2}
\end{algorithm}

\paragraph{Selection Strategy} We used \emph{softmax selection} strategy. We used same strategy as for meta-solver for adding crossover individuals and to maintain the size of population above $20\%$ of population in first generation.

\begin{table}[h]
    \caption{Hyperparameters for TSP Solver}
    \label{tab:tsp}
    \centering
    \begin{tabular}{|c|c|}
    \hline
    \textbf{Hyperparameter} & \textbf{Value}\\
    \hline
     Initial Population Size    &  200\\
     Minimum population Size    & 50\\
     Mutation Rate & 0.02\\
     Crossover rate & 0.7\\
     Point crossover probability & 0.5\\
         \hline
    \end{tabular}

\end{table}

\subsection{Results}
We found that meta-solver was very sensitive to hyperparameters. The hyperparameters are listed in Table \ref{tab:mtsp} and \ref{tab:tsp}.

\begin{figure}[h]
        \centering
        \includegraphics[width=.7\textwidth]{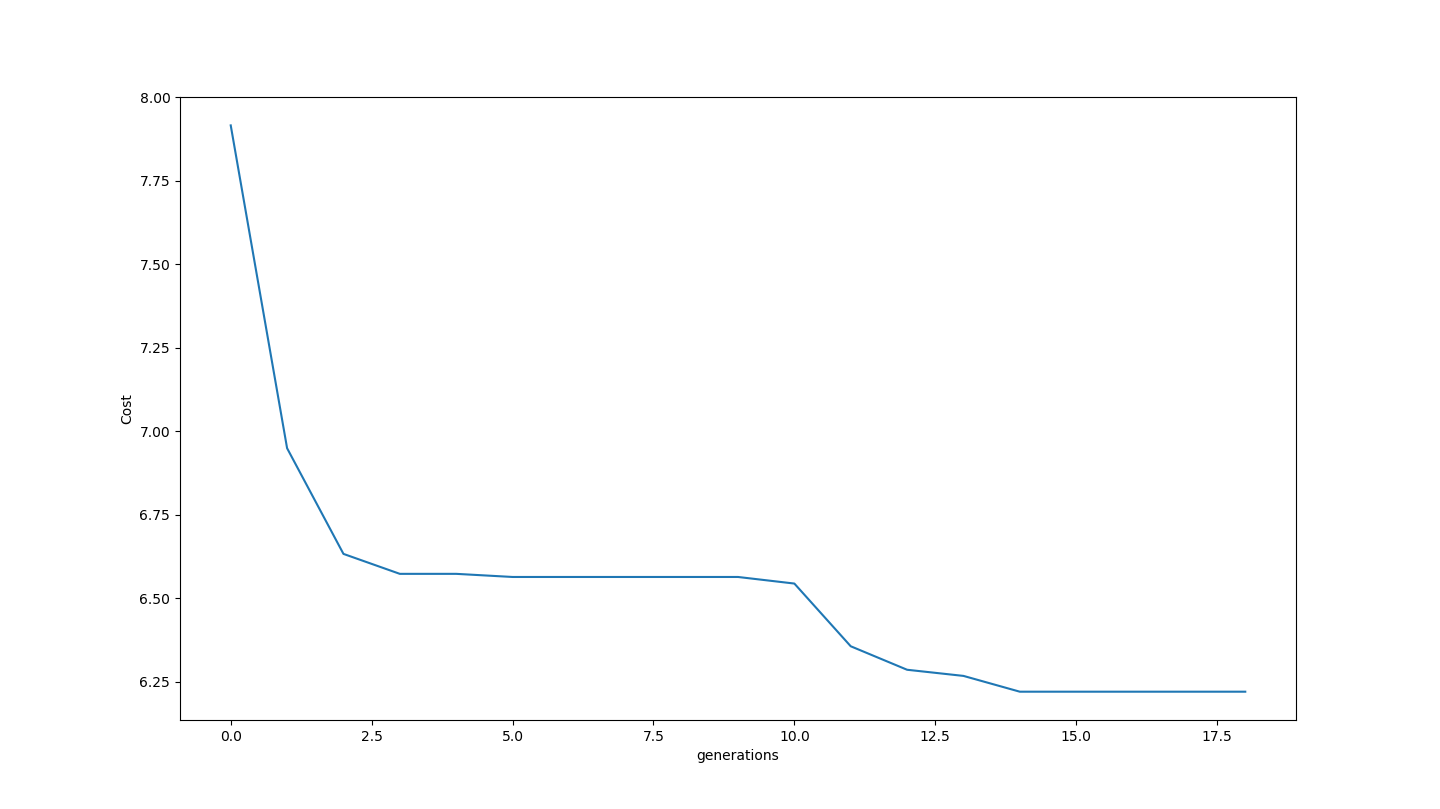}
        \caption{Cost curve for 30 vertex graph}
        \label{fig:tsp1}
    \end{figure}

First we ran the algorithm on $30$ vertex graph generated by sampling points in $[0,1]\times[0,1]$. We set the penalty for each vertex to be uniform and equal to $0.4$. The learning curve is shown in Figure \ref{fig:tsp1}. The $2-$approximate greedy solution by \cite{williamson2011design} was bettered within $2$ generations.

\subsection{Adaptation to change in constraints}
We experimented to see how the meta-solver adapts to change in penalty values.

\begin{figure}[h]
        \centering
        \includegraphics[width=.9\textwidth]{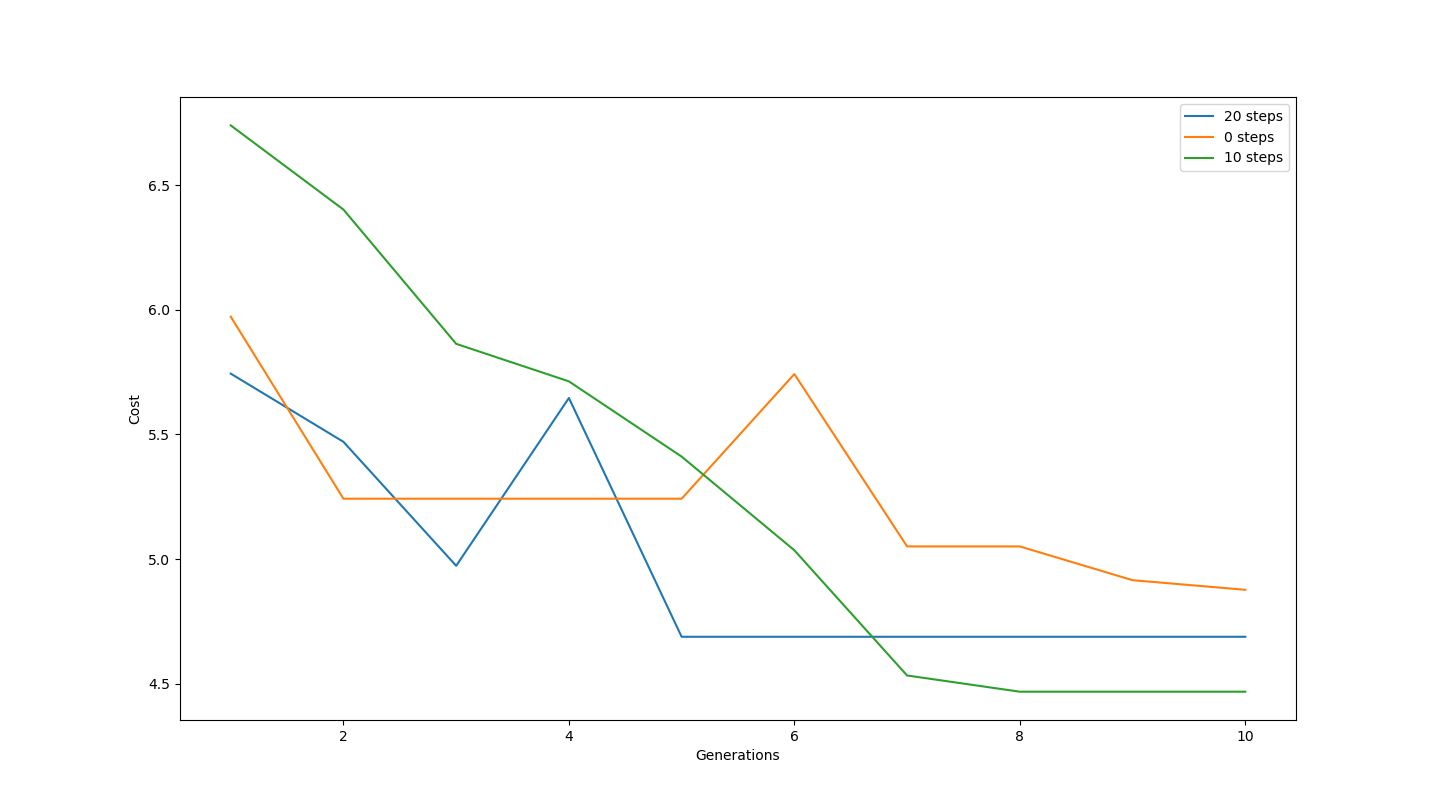}
        \caption{Learning curve depending $t$, number of generation the previous problem was trained for}
        \label{fig:tsp_comp}
\end{figure}

We first selected $20$ random vertices and set their penalty very high $(=10)$ and rest very low $(=0.1)$ and trained for $t$ generations. Then, we chose another $20$ random vertices and made their penalty high with others having low penalty. If $t$ is too low, it is similar to to starting from scratch. If $t$ is too high (around 20) the available solutions are not diverse enough to quickly adapt to changed constraint in further iterations. The comparison is visualized in Figure \ref{fig:tsp_comp}.

\paragraph{Training with gradual change in constraints}
At suitable value of $t$, the solutions contained subsets of points that have low cost path solutions. Hence, it easily adapted to change in constraints. Using this insight, we hypothesized that rather than giving the actual penalties for meta-solver from first iteration, in which case the meta-solver only optimizes to prune the size of vertex set, we can initially have a high penalty so that meta-solver learns good local solutions from different parts of the graph. Then we slowly transform the penalty term to get it closer to actual penalty.
The procedure is described in Algorithm \ref{alg:adapt_tsp}.

\begin{algorithm}[h]
\SetKwInOut{Input}{input}
\SetKwInOut{Output}{output}
\SetAlgoLined
\Input{penalty array $P[1\dots|V|]$}
Initialize meta-solver $M$\;
Initialize array $P'$ with all values equal to $\max_i P[i]$\;
\For{$i\leftarrow[|V|]$}{
    $diff[i]\leftarrow \frac{(P'[i]-P[i])}{n/2}$
}

\For{$i\leftarrow\{1,\dots,n\}$}{
    Input $P'$ as penalty for $M$\;
    Train $M$ for 2 generations\;
    $\forall i\in[|V|]: P'[i]\leftarrow P'[i]-diff[i]$\;
}
Input $P$ as penalty for $M$\;
Train $M$ till convergence\;

\caption{Adaptive training for TSP solver}
\label{alg:adapt_tsp}
\end{algorithm}

\begin{figure}[h]
    \centering
    \includegraphics[width=\textwidth]{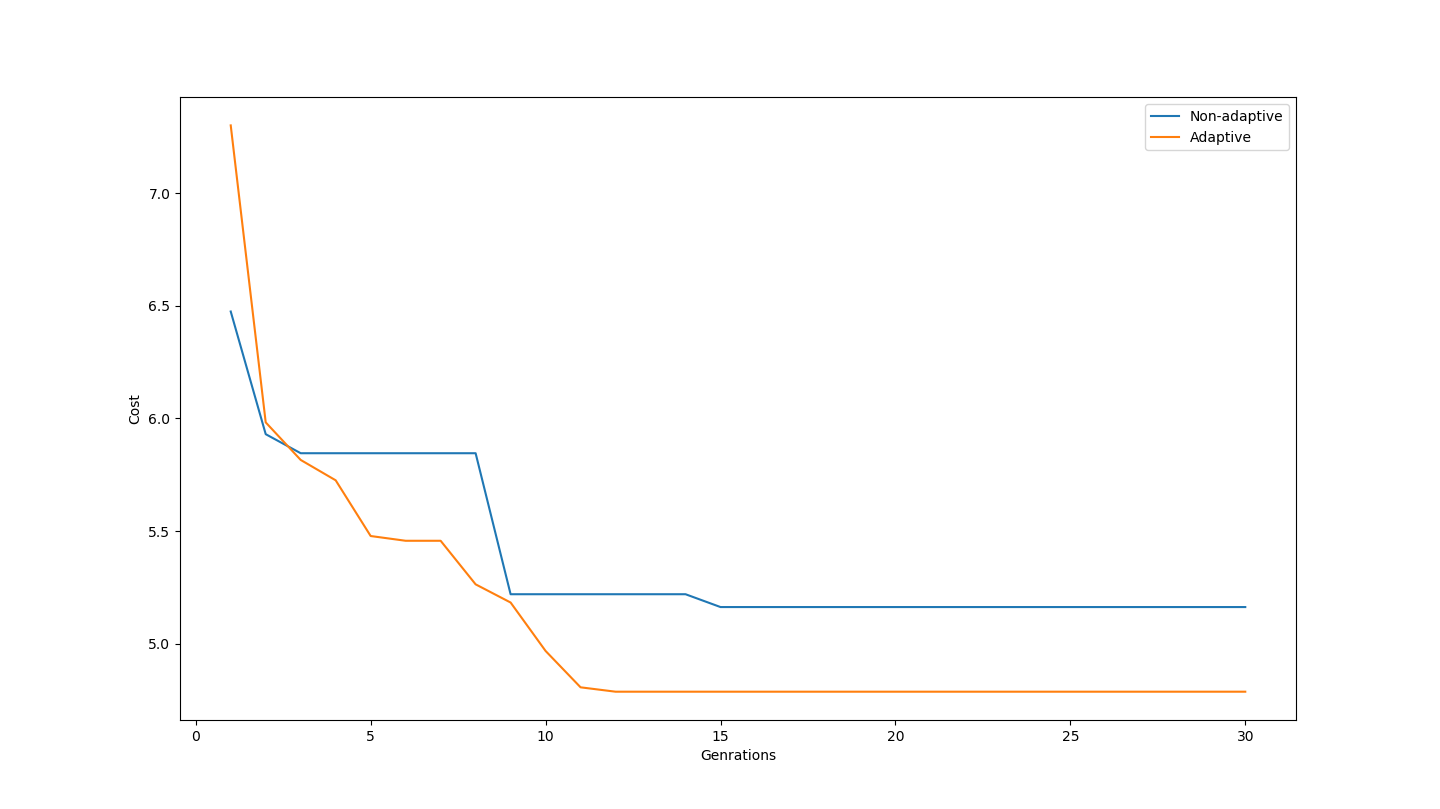}
    \caption{Adaptive vs Non-adaptive training}
    \label{fig:adapt}
\end{figure}

We experimented this strategy for $30$ vertex graph problem with penalties in range $0.25 \pm 0.2$. The comparison between adaptive and non-adaptive strategies is visualized in Figure \ref{fig:adapt}. Initially the cost of solutions were very high. However, as the penalties became closer to actual penalties, we converged to a better solution than when penalties remained fixed.

\section{Regression with unknown objective function}
\paragraph{Motivation} In real world, when we are solving a complex task, the exact nature of the reward or objective function maybe hard to decipher. We use many well known objectives with relative importance measures try and increase the reward signal or mitigate loss. For example, how fit an individual of a species is to propogate its genetic information to next generation may depend on many objective functions which are simpler to define, such as resistance to diseases, mental prowess, physical strength, social skills, fertility, etc. The overall objective function maybe even more complicated and change from time to time. Hence, population focuses on optimizing these simpler objectives with varying importance which evolves with time.

\paragraph{Problem Statement}
Given input data $\{(x_1,y_1),(x_2,y_2),\dots,(x_n,y_n)\}$, polynomial regression problem involves finding a polynomial $p:\mathbf{R}\to \mathbf{R}$ that best approximates $y_i$ for each $x_i$. Regression task can be solved using any of the objective functions described in Section 2.3.

Now, assume we don't know what objective function we need to minimize but we have an oracle that on receiving the predicted values of $y$ gives out a cost signal which we need to minimize. The GA agent need to find the best combination of available objective functions to minimize the error. In particular the objective function is of the form
\[MSE\_error+\lambda_1Quantile\_error(\gamma)+\lambda_2L2\ regularization\]
and the agent also assumes the degree of polynomial to be $d$. So, the meta-solver needs to search the right set of hyperparameters $(\lambda_1, \lambda_2,d,\gamma)$ for which individual GA solvers find the right coefficients of polynomial.

\subsection{Setup}
The input data is generated from the function $y=p(x)+\epsilon$ where $p$ is a polynomial and $\epsilon$ is Gaussian noise added.
The GA solvers have fixed hyperparameters $(\lambda_1, \lambda_2,d,\gamma)$ and use GA to find the coefficients of polynomial $p(x)=\sum_{i=0}^d a_ix^i$. The meta-solver searches for right set of hyper-parameters.
\subsection{Details of meta-solver}
\paragraph{Genotype} Each individual $i$ has values for hyper-parameters in array $H_i = (\lambda_1, \lambda_2,d, \gamma)$.

\paragraph{Crossover} The algorithms for crossover is same as Algorithm \ref{alg:co_tsp} where the the values of arrays are swapped with some probability $c$.

\paragraph{Mutation} We perturb the values in hyper-parameter array of individuals. $\lambda_1 ,\lambda_2$ and $d$ are perturbed by adding a noise from normal distribution with standard deviation 0.5 and 0.1 respectively. Degree is changed by adding or subtracting 1 (when $d>0$).

\paragraph{Selection Strategy} We use the \emph{percentile selection} strategy and same heuristic as used for meta-solver of TSP.
\begin{table}[h]
    \caption{Hyperparameters for meta-Solver for Regression}
    \label{tab:mreg}
    \centering
    \begin{tabular}{|c|c|}
    \hline
    \textbf{Hyperparameter} & \textbf{Value}\\
    \hline
     Initial Population Size    &  100\\
     Minimum population Size    & 20\\
     Mutation Rate & 0.2\\
     Crossover rate & 0.5\\
     Point crossover probability & 0.5\\
     Number of generations of sub-solvers for one generation of meta-solver & 200\\
         \hline
    \end{tabular}

\end{table}

\subsection{Details of GA solver for regression}
Consider a GA regression solver for degree $d$ polynomial which approximates $p(x)=\sum_{i=0}^d a_ix^i$.
\paragraph{Genotype} The weights of individual $i$, $A_i=\{a_{i,1}, a_{i,2}, \dots, a_{i,d}\}$.
\paragraph{Crossover} The algorithms for crossover is same as Algorithm \ref{alg:co_tsp} where the the values of arrays $A_i,A_j$ are swapped with some probability $c$.

\paragraph{Mutation} We perturb the weights by adding noise from normal distribution $\mathcal{N}(0,2)$.

\paragraph{Selection Strategy} We use \emph{random tournament selection} to select half the previous generation before doing mutation and crossover.

\begin{table}[h]
    \caption{Hyperparameters for Polynomial Regression solver}
    \label{tab:reg}
    \centering
    \begin{tabular}{|c|c|}
    \hline
    \textbf{Hyperparameter} & \textbf{Value}\\
    \hline
     Initial Population Size    &  500\\
     Minimum population Size    & 100\\
     Mutation Rate & 0.2\\
     Crossover rate & 0.7\\
     Point crossover probability & 0.5\\
         \hline
    \end{tabular}

\end{table}

\subsection{Results}
We fixed the the target polynomial function $p(x)$ as
\begin{equation}
    p(x)=4x^2 + 3x + 4 + \epsilon, \epsilon\leftarrow\mathcal{N}(0,0.2)
\end{equation}

\begin{figure}[h]
    \centering
    \includegraphics[width=.8\textwidth]{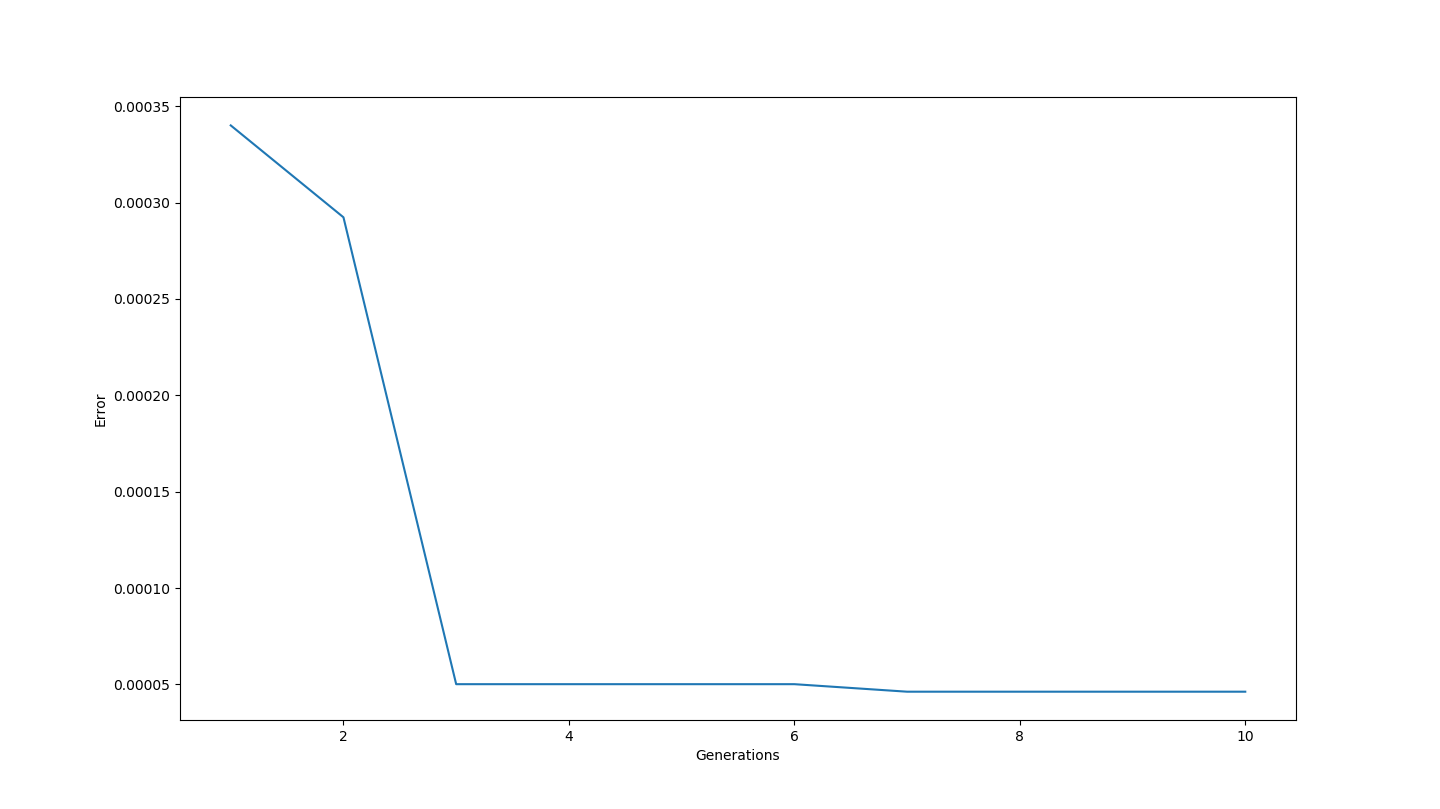}
    \caption{Learning curve for degree 2 polynomial}
    \label{fig:reg_1}
\end{figure}

The target objective function to solve if \emph{Huber loss}(see Section 2.3) with $\delta=0.2$. The learning curve is in Figure \ref{fig:reg_1}.

\begin{figure}[h]
    \centering
    \includegraphics[width=.8\textwidth]{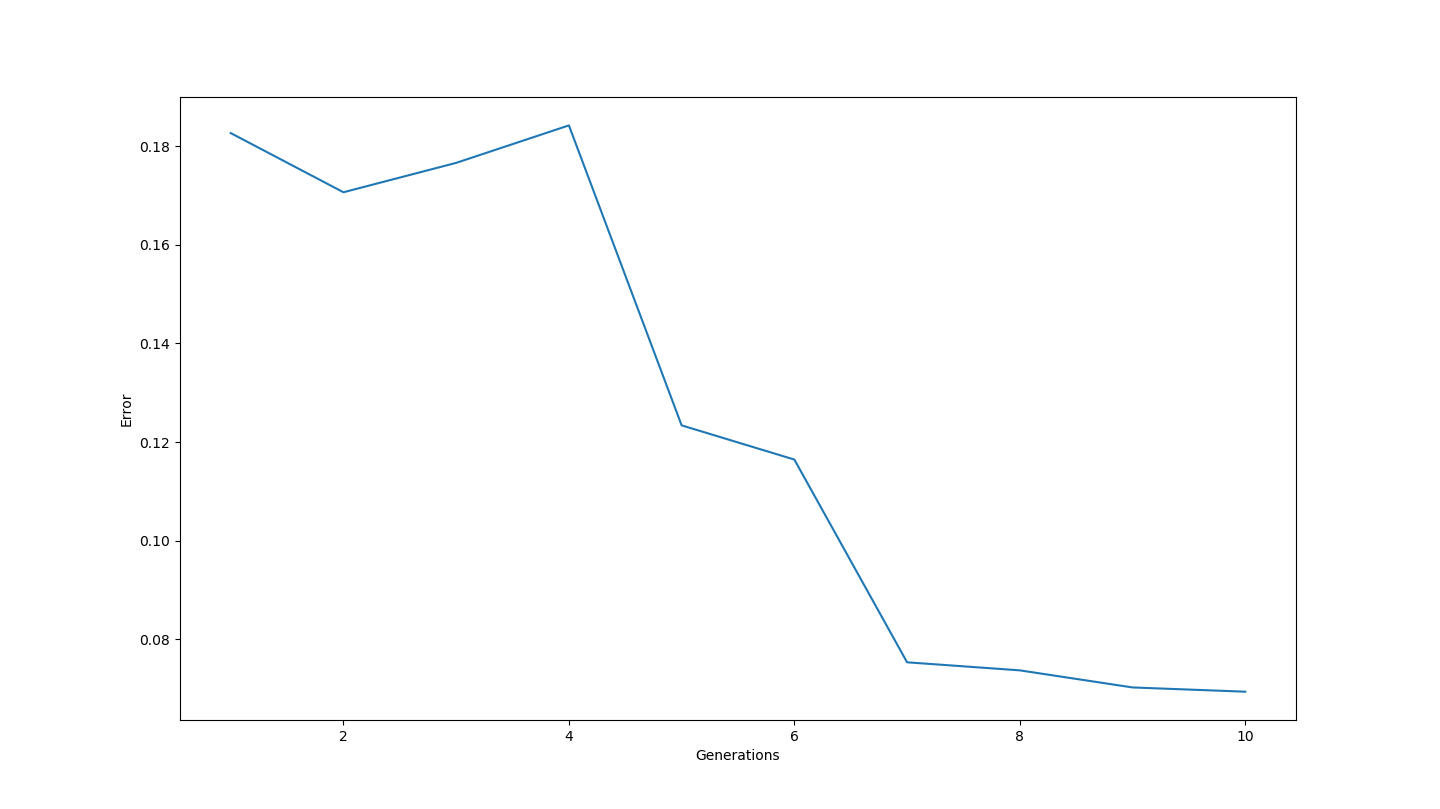}
    \caption{Learning curve for degree 6 polynomial}
    \label{fig:reg_2}
\end{figure}

\begin{figure}[h]
    \centering
    \includegraphics[width=.8\textwidth]{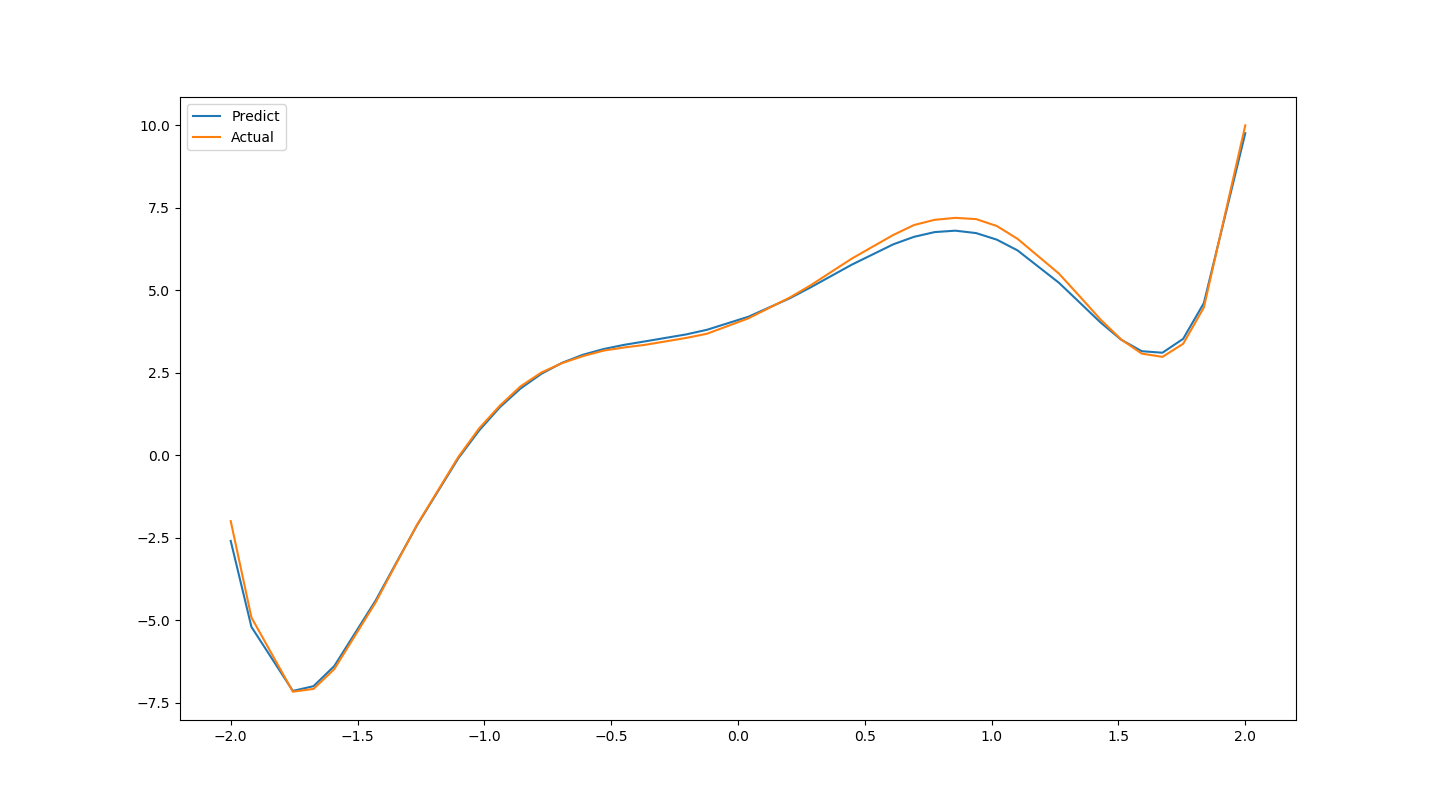}
    \caption{Predicted function for degree 6 polynomial}
    \label{fig:reg_3}
\end{figure}

\begin{figure}[h]
    \centering
    \includegraphics[width=.8\textwidth]{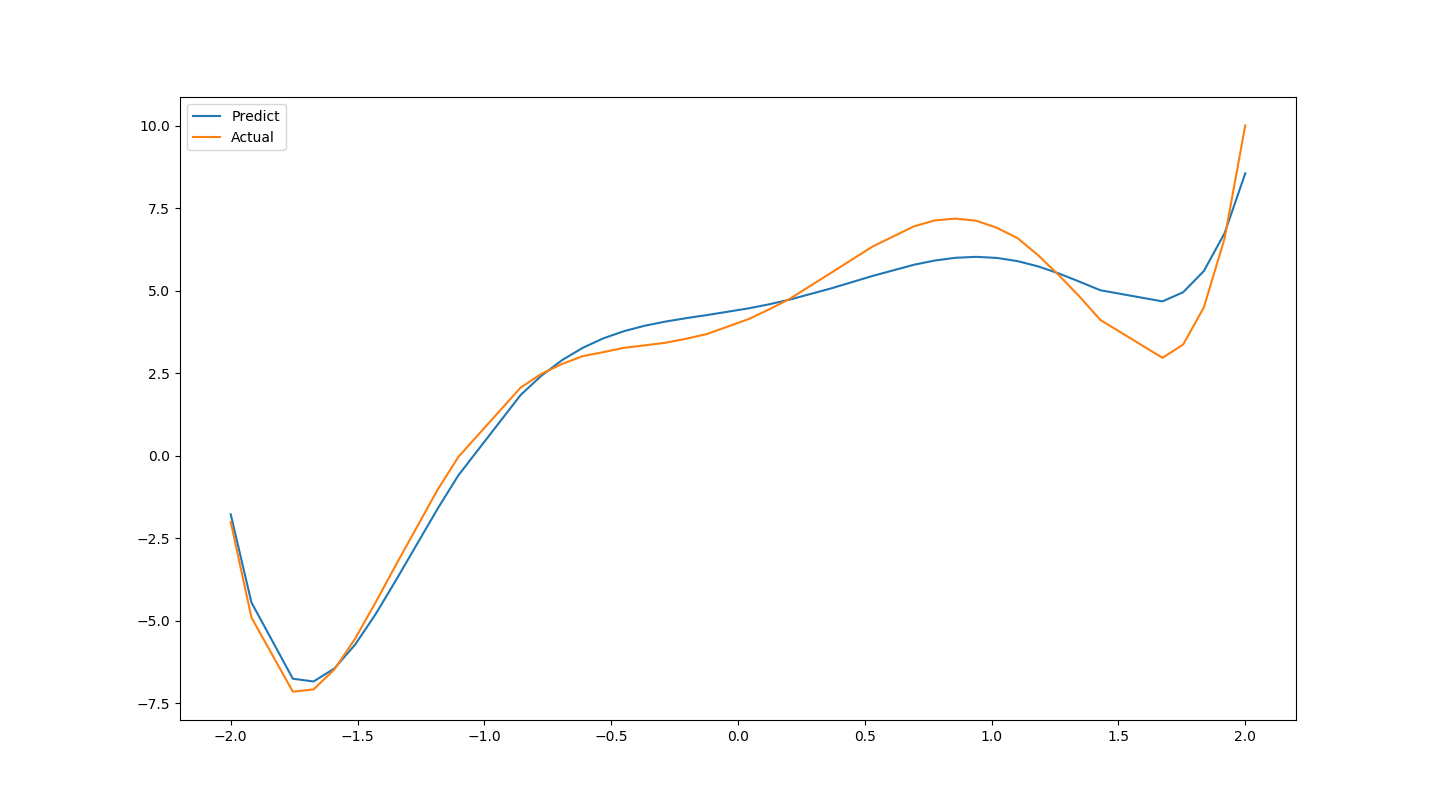}
    \caption{Predicted function in generation 2 for degree 6 polynomial with unweighted loss function}
    \label{fig:reg_6}
\end{figure}

We then increased the degree of polynomial to 6.
\begin{equation}
    p(x)=x^6-5x^4 + 4x^2 + 3x + 4 + \epsilon, \epsilon\leftarrow\mathcal{N}(0,0.2)
\end{equation}
It took slightly longer to converge as seen in Figure \ref{fig:reg_2}. The resultant solution was actually a degree 7 polynomial(see Figure \ref{fig:reg_3}).
We saw that in first few generations for $x>0$, the curve didn't approximate actual function very well. Hence, we decided do test with an adapted training method.(See figure \ref{fig:reg_6})

\begin{figure}[h]
    \centering
    \includegraphics[width=.8\textwidth]{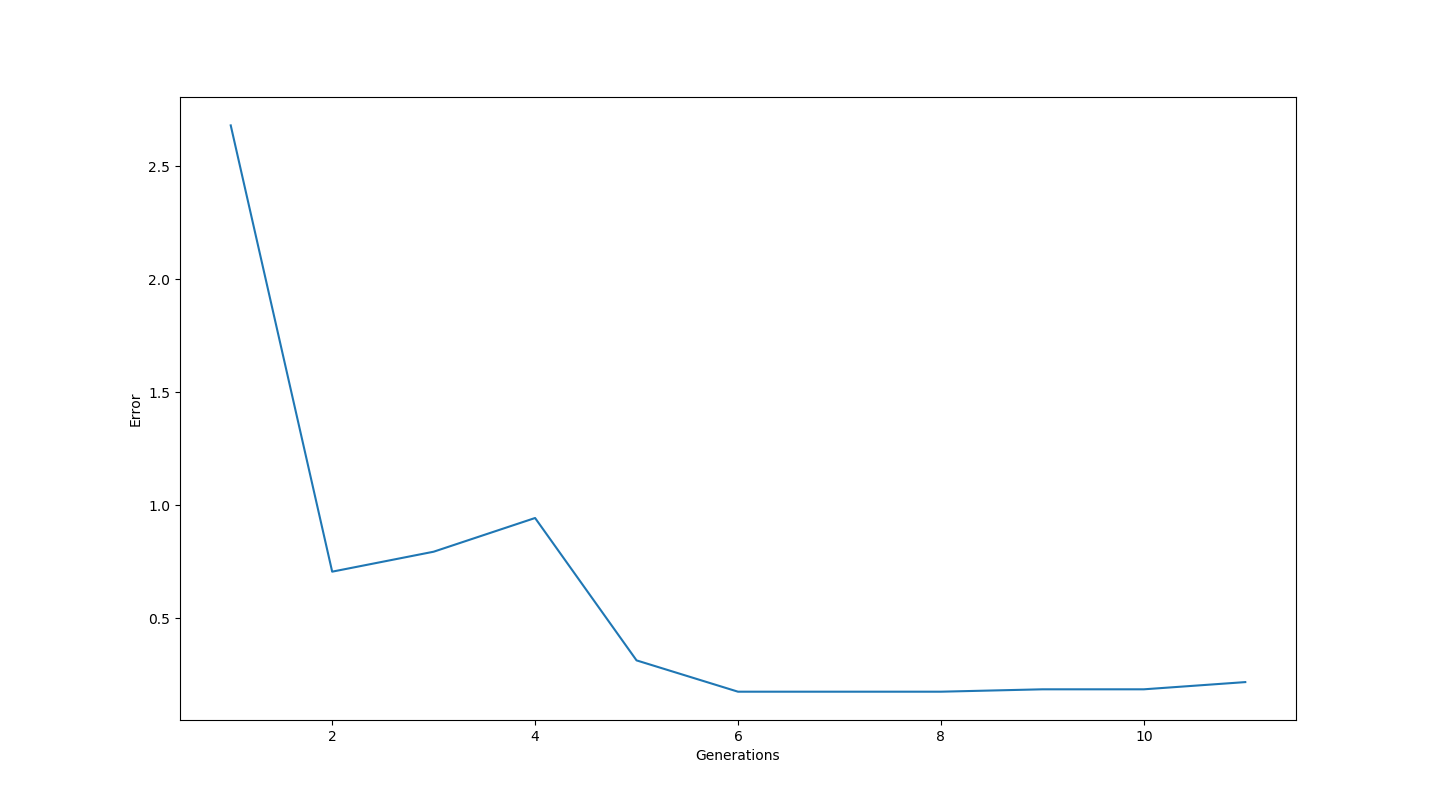}
    \caption{Learning curve for degree 6 polynomial with weighted loss function}
    \label{fig:reg_5}
\end{figure}

\begin{figure}[h]
    \centering
    \includegraphics[width=.8\textwidth]{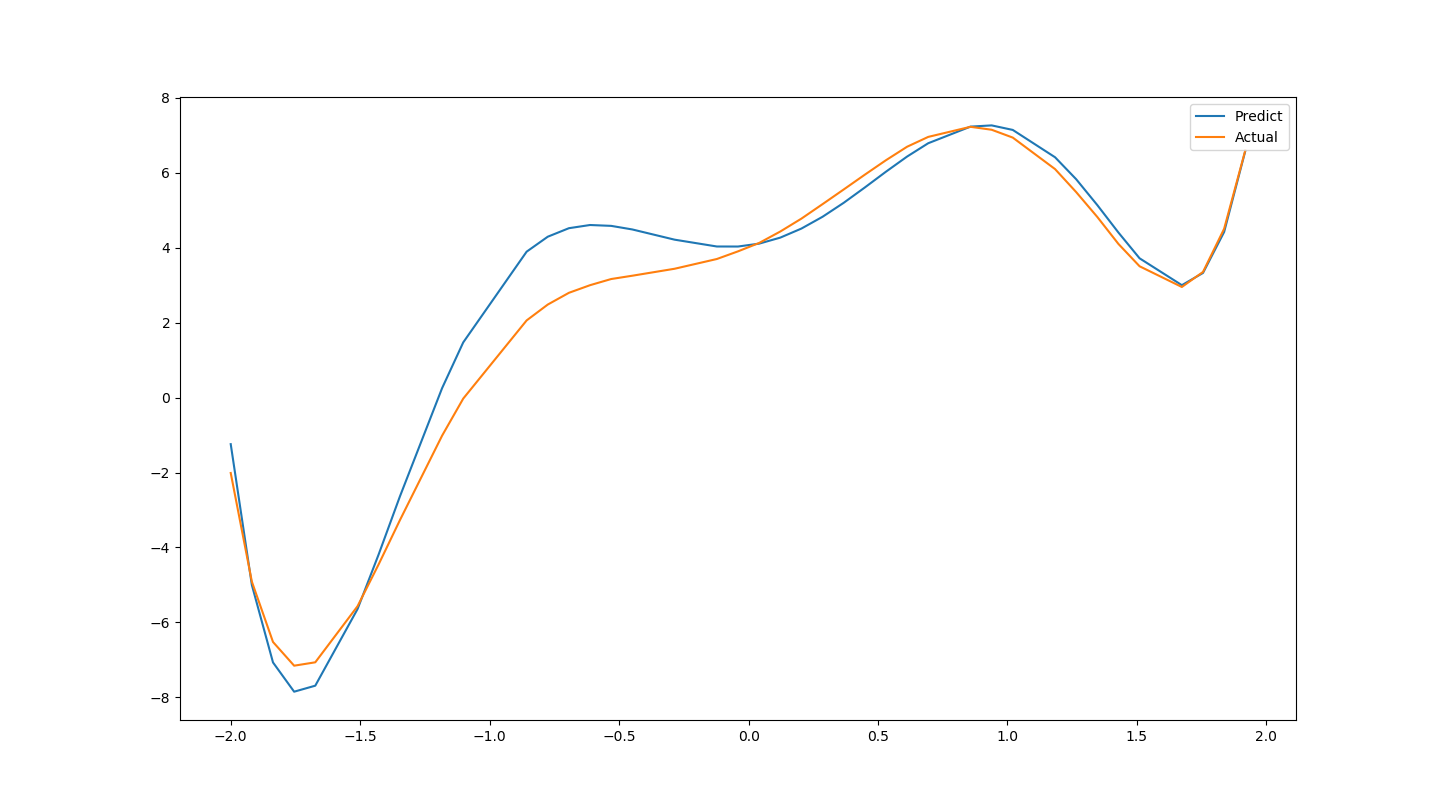}
    \caption{Predicted function in generation 2 for degree 6 polynomial with weighted loss function}
    \label{fig:reg_4}
\end{figure}

\paragraph{Adapting to weighted objective Function}
Next we tested how the solver would respond to different weights for different ranges of $x$.
We multiplied the loss term by 10 for all terms with corresponding $x>0$ to check how the solver reacts to different constraint. In first few generations, it focused only on value for $x>0$ and approximated a degree 4 polynomial and quantile parameter $\gamma$ was high, (see Figure \ref{fig:reg_4}) then later it approximated entire function while decreasing $\gamma$ value. (See the ridge in loss curve in Figure \ref{fig:reg_5}) We found that this training procedure converge faster than with usual huber loss. This is because, the curve at $x>0$ shows large variance and may be harder to approximate while left side is easier.

\clearpage
\section{Discussion}
We have used our hierarchical genetic algorithm framework to solve the \texttt{soft-TSP} and polynomial regression. In case of \texttt{soft-TSP} it beat the 2-approximate greedy solution within two generations by large margin. In case of polynomial regression, we used combination of different objective functions to approximate hidden objective function. Based on the hidden objective function, the best solution adapted to minimize the hidden objective faster. Based on observations during running the code we found that compute time for the solving uni-variate polynomial regression problem using our framework took much less time than using gradient descent algorithms. This may not the case if the target function is complex and multivariate.

In both the problems, the key challenge was to maintain the diversity of the solutions so that the meta-solver doesn't stagnate to a locally optimal objective function. Hence, we set mutation rate and crossover rate very high for meta-solver and also used mutation to maintain a large population at every iteration. Another challenge was to determine number of generations of lower level solvers to spawn before evolving the objective function. If evolution of objective functions happens too fast, we may not be able to access the \emph{fitness} of the objective function in approximating the actual objective.

This phenomenon of maintaining diversity is similar to how nature maintains wide genetic variation in species even when a very small group of species are much fitter at given time by employing various strategies so that different groups individuals have higher score for difference measures of competence for survival \cite{tabachnick2013nature} and have mechanisms for co-existence in different ecological conditions\cite{levine2009importance}.

\section{Future work}

We can test our framework in other kinds of optimization problems where a objective function can be solved by searching over a family of simpler objective functions. This is straightforward in cases where there exists notions of trade-off between parameters. Fixing one set of parameters at one level and solving for other set of parameters is easily adapted to this framework.

One can also explore how, during the mutation and crossover of objective functions, the solutions of these objective functions can be adapted to offspring objective function. This may enable further improvement in compute time.

%
%

%
%
%
 \bibliographystyle{splncs04}
 \bibliography{references}
%
\end{document}